# HADES: Hardware Accelerated Decoding for Efficient Speculation in Large Language Models

*Ze Yang[1st & *], Yihong Jin[2nd], Xinhe Xu[3rd]*
*University of Illinois Urbana-Champaign*
*Champaign, USA*
*{zeyang2, yihongj3, xinhexu2}@illinois.edu*

***Abstract:*** **Large Language Models (LLMs) have revolutionized natural language processing by understanding and generating human-like text. However, the increasing demand for more sophisticated LLMs presents significant computational challenges due to their scale and complexity. This paper introduces Hardware Accelerated Decoding (HADES), a novel approach to enhance the performance and energy efficiency of LLMs. We address the design of an LLM accelerator with hardware-level speculative decoding support, a concept not previously explored in existing literature. Our work demonstrates how speculative decoding can significantly improve the efficiency of LLM operations, paving the way for more advanced and practical applications of these models.**

***Keywords: Large Language Model (LLM); Hardware Accelerators; FPGAs; GPU; Transformer; Energy Efficiency; Speculative Decoding***

## I. Introduction

Large language models (LLMs) have gained a lot of attraction in recent years, revolutionizing the way we interact with technology and process information. Their impact spans a wide range of applications, from natural language processing, content creation, and translation services to more complex tasks like programming assistance, data analysis, and even aiding in research and discovery across various fields.

As the demand for LLMs increases, the need for a higher token generation rate also increases drastically. Agents driven by LLM use techniques such as Chain-Of-Thought [23, 24] to accomplish complex tasks autonomously. In these cases, the token generation rate directly affects the execution time of the autonomous agents.

Building on this momentum, our work delves into the exploration and implementation of Hardware-Accelerated Speculative Decoding. This innovative approach aims to significantly elevate the token generation rate, addressing the growing demand for efficiency in LLM applications. By leveraging hardware acceleration, we seek to enhance the speed and responsiveness of LLM-driven agents, making complex tasks more feasible in real-time scenarios. Our initiative marks a pivotal step towards optimizing the performance of large language models, ensuring they can meet the needs of advanced applications and deliver results with unprecedented speed.

This paper makes the following contributions:

1. We devise a modular design that enables speculative decoding in hardware with simply an add-on to the original ASIC implementation.

2. We demonstrated that by implementing speculative decoding pipelines in hardware, significant speedup and power efficiency improvements can be achieved.

## II. Background and Motivation

With the rise of popularity of AI in language generation, Generative Pre-trained Transformers have seen an exponential growth in their size, or weights. As the size for autoregressive models [17, 19, 20, 25] increases, so do their capabilities as demonstrated by models such as LLaMa [15, 18] and GPT-3 [1]. However, with the increase in size, comes at a cost of inference time. Since a standard inference pass requires all weights in the model to be used at least once, leading to most large models being memory bottlenecked [15]. Moreover, each decoding step is dependent on the previous step and therefore must be done sequentially. This leads to inefficiency on GPUs, which are optimized for parallel workload, that often see utilization in the single-digits.

Increasing batch size allows the model to process multiple input streams in parallel and allows for efficient utilization and throughput, with slightly worse latency. However, batching is only useful in a serving environment where multiple users are calling the GPT at the same time. Most of the GPT usage is at a personal level, where the batch size is 1. There have been various attempts to reduce the runtime for a single decoding step by either reducing the amount of computation for all workload, such as quantization [5], or ones that are observed to be easily generated, such as early-exit [14]. While these methods have proven successful in reducing the time for decoding steps, they either require modification to the model architecture or do not maintain the same probability distribution as the original model. Speculative decoding [7, 21, 26], on the other hand, does not require modification to the original architecture and guarantees the same probability distribution. Speculative decoding aims to dynamically reduce the workload for a large autoregressive model by utilizing a smaller autoregressive model for most of the decoding steps. This is under the observation that some decoding steps are much easier than others, and that the size of the model makes minimal impact on the probability distribution.

Speculative decoding aims to reduce the runtime for a given model, which we will refer to as the target model, by employing a smaller model using the same encoder and tokenizer, which we will refer to as the draft model. After tokenizing and encoding the input, the prompt tokens will be fed through a series of autoregressive runs as follows. The draft model will run $\gamma$ times autoregressively, generating $\gamma$ speculated tokens. The target model will then run once, verifying the speculated tokens and generating the $\gamma+1^{th}$ token in parallel. Each speculated token is then examined and accepted if the said token could be generated by the target model. If all speculated tokens are accepted, the speculated tokens as well as the $\gamma+1^{th}$ token will be appended to the prompt tokens for the next series of runs. Observe that the runtime reduction is achieved with lower compute and memory utilization. Consider the ideal case where all speculated tokens are accepted, the series of runs have effectively generated $\gamma+1^{th}$ tokens in $\gamma * t_{draft} + t_{target}$, where the naive autoregressive runs on the target model only will take $(\gamma+1) * t_{target}$. Since the draft model is smaller than the target model, then $t_{draft} < t_{target}$. If there is a mispredict, the algorithm accepts all speculated tokens before the mispredicted one and restarts the series of runs.

Speculative decoding drew inspiration from speculative execution [2], where tasks are being performed speculatively in parallel to the verification for the necessity of said task. Branch prediction is a prominent example for speculative execution, where instructions are fetched and executed under the assumption that they are correct. Similarly, speculative decoding runs the smaller model speculatively under the assumption that the decoding steps are mostly model size agnostic.

Although the concept of speculative decoding has only existed for roughly a year, there have been numerous advances in the software area, from further optimization with the choice of draft models to the techniques using draft-less models. This proves the genuineness and the usefulness of the concept of speculative decoding. Due to the short time span since discovery and speculative decoding's origin as a software optimization technique, there has been minimal discourse in the hardware aspect. Therefore, we believe there exists an opportunity for hardware support of speculative decoding. As mentioned above, speculative decoding does not require architectural modification to the original model. We believe that our accelerator should adhere to that philosophy, where our accelerator does not require changes to the accelerator for the model itself, if one were to choose to integrate it. This significantly limits our design space as we need to consider consistency between different methods of integration, between hardware and software, as well as different hardware accelerators.

## III. DESIGN AND IMPLEMENTATION

To explore the impact of hardware-accelerated speculative decoding, we initially conduct an in-depth analysis of serving OPT[11] and Llama[13] , the most widely utilized large language models, utilizing software-based speculative decoding. Subsequently, we compare these findings with those obtained through our approach of hardware-accelerated speculative decoding.

**Setup**. For software-based speculative decoding, we will implement in the Microsoft DeepSpeed library [12]. We use the same setup as SpecInfer [10], first using the draft model to generate draft tokens and then using the target model to verify the output of the draft model. We set the batch size to 1 and use greedy decoding. We perform both software-based or accelerated by hardware experiments on the OPT model and GPT-2 base models. For ASIC implementation of speculative decoding, we will use Synopsys VCS to compile HLS code from PyTorch models. We will then add the proposed speculative decoding modules and simulate them with VCS. The synthesized power and token generation rate will be recorded.

**Metrics**. To evaluate the efficacy of various speculative decoding approaches in the context of a target model, we focus on two key metrics: throughput, defined as the number of tokens generated per second, and the Token Acceptance Rate (TAR). It's important to highlight that the main objective behind implementing speculative decoding strategies, whether software-based or accelerated by hardware, is to enhance the overall throughput.

**Estimated output**. In the anticipated outcomes of our comparative analysis, we project that hardware-accelerated speculative decoding will demonstrate a superior throughput relative to its software-based counterpart. This expectation is rooted in the inherent efficiency gains that hardware acceleration brings to computational tasks, allowing for more rapid data processing and token generation. Such improvements are critical in environments where the speed of inference plays a pivotal role in the overall performance and responsiveness of the system. Hence, we predict a notable enhancement in throughput metrics for hardware-accelerated methods, underscoring their potential to optimize speculative decoding processes significantly.

**Challenges**. Since implementing an end-to-end LLM accelerator is very complicated and unnecessarily complex, we decided to use custom logic to speed up only the verification part of speculative decoding. This way, our logic will be compatible with all existing LLM accelerators, and it will greatly speed up the development. We first try to use OpenHLS to lower Python to RTL using HLS techniques. However, due to the lack of maintenance in the codebase, we cannot successfully translate it. Next, we wrote a C implementation and then used AMD Vivado to lower C into RTL. It does the translation, but the testbench suggests that the performance is very poor. Then we looked at the HLS RTL carefully and optimized its performance by writing our own Verilog code shown in figure 1.

This results in significant speedup compared to the GPU baselines, see section 5 for more discussion.

## IV. EXPERIMENTS

We initially planned to utilize tools like OpenHLS [8] and scalehls [4] to convert a PyTorch model into an RTL flow, initializing draft and target models for speculative decoding on FPGA. PyTorch Labs introduced gpt-fast [6], an efficient, PyTorch-native transformer optimization framework designed specifically for large language models. gpt-fast requires no dependencies other than PyTorch and SentencePiece and supports both INT8/INT4 quantization and speculative decoding, aligning perfectly with our objectives. Importantly, the LLaMA family is also supported by gpt-fast. Therefore, we commenced optimization of the Llama-2-7b-chat-hf and Llama-2-70b-chat-hf models using gpt-fast, addressing some PyTorch version compatibility issues in the process. We successfully completed the INT8 quantization and speculative decoding optimizations.

However, we encountered two significant challenges. First, gpt-fast uses a nightly version of PyTorch that incorporates many new features unsupported by our chosen RTL conversion tool. Second, the optimized models remain excessively large for effective FPGA simulation; for instance, the INT8 version of Llama-2-7b-chat-hf still requires 6.5GB, severely slowing our simulations.

Due to these challenges and the limited timeframe, we decided to pivot from the LLaMA models to explore alternatives like GPT-2 and OPT.

TABLE I. PERFORMANCE CHART

| Model Pair | Scheme | Token/sec | Speedup |
|---|---|---|---|
| GPT2<br>Draft = 124M<br>Target = 1554M | $\gamma = 0$ | 77 | 1.0x |
| | $\gamma = 2$ | 102 | **1.32x** |
| | $\gamma = 4$ | 95 | 1.23x |
| | $\gamma = 8$ | 84 | 1.09x |
| | $\gamma = 16$ | 51 | 0.66x |
| OPT<br>Draft = 125M<br>Target = 6.7B | $\gamma = 0$ | 38 | 1.0x |
| | $\gamma = 2$ | 56 | 1.47x |
| | $\gamma = 4$ | 70 | **1.84x** |
| | $\gamma = 8$ | 60 | 1.58x |
| | $\gamma = 16$ | 34 | 0.89x |

We conducted experiments with both the GPT-2 and OPT model families. We selected these because of their widespread use and extensive support on platforms like Huggingface, which facilitates troubleshooting. These model families offer a broad spectrum of sizes, ranging from 100 million to 60 billion parameters. This variety provides us with greater flexibility in our hardware development efforts.

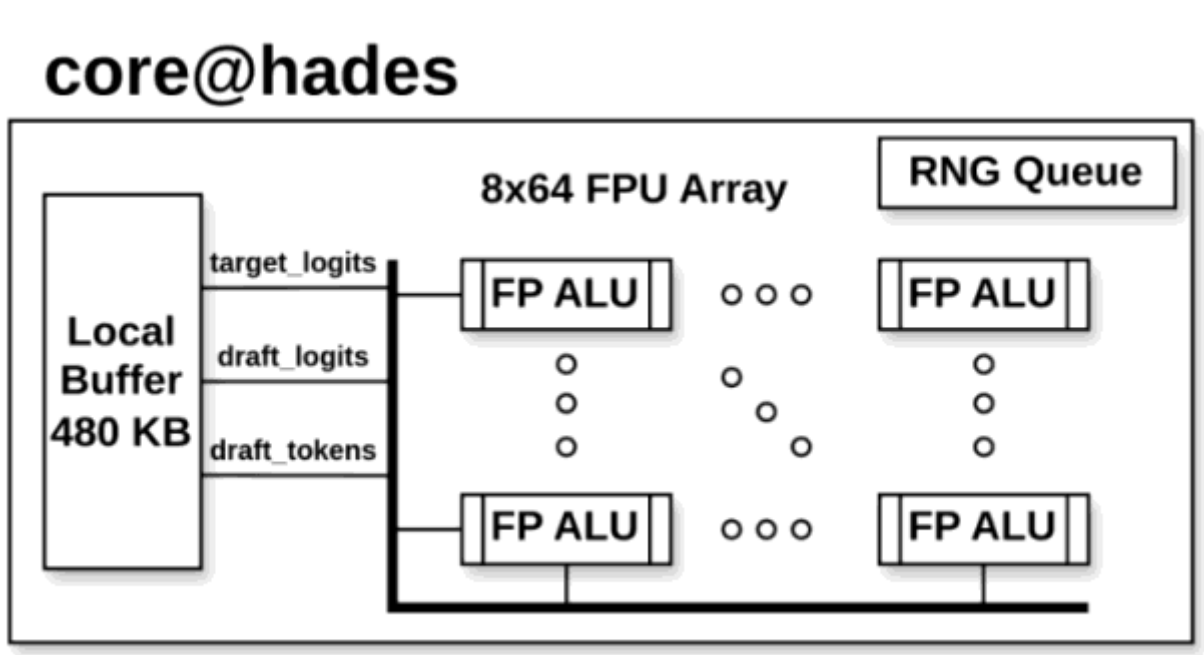

Figure 1. *HADES architecture diagram.*

Utilizing Torch MLIR, we are able to lower PyTorch model into MLIR, however, we encountered significant challenges when lowering from MLIR to Verilog. Alternatively, we are actively exploring the probability of using ScaleHLS to lower MLIR to HLS C++, and then from HLS C++ to RTL using AMD Vitis software. We implemented Algorithm 1. From Chen et al. [3].

Algorithm 1 Speculative Decoding

1. Given $\gamma$ and final sequence length T
2. Given draft model p, target model q and initial tokens $x_0, ..., x_t$
3. $n \leftarrow t$
4. While $n < T$ do:
5. For $t \in \gamma$ :
6. Decode draft model autoregressively:
7. $\tilde{x}_t \sim p(x \mid x_1, ..., x_n, \tilde{x}_1, ..., \tilde{x}_{t-1})$
8. end for
9. Decode target model:
10. $q(x \mid x_1, ..., x_n), ..., q(x \mid x_1, ..., x_n, \tilde{x}_1, ..., \tilde{x}_t)$
10. For $t \in \gamma$ do
11. Sample $r \sim U[0, 1]$ (uniform distribution)
12. If $r < \min(1, q(x \mid x_1, ..., x_{n+t-1}) / p(x \mid x_1, ..., x_{n+t-1}))$:
13. $x_{n+t} \leftarrow \tilde{x}_t$
14. $n \leftarrow n + 1$
15. else:
16. Sample $x_{n+t} \sim q(x \mid x_1, ..., x_{n+t-1}) - p(x \mid x_1, ..., x_{n+t-1})$
17. break
18. end if
19. end for
20. If all $\gamma$ accepted:
21. Sample $x_{n+t+1} \sim q(x \mid x_1, ..., x_{n+t})$
20. $n \leftarrow n + 1$
22. end if
23. end while

In the experiment, table 1, the baseline is the performance of the target model when running autoregressively. We then apply speculative decoding with varying gamma factors and observe how the performances changes.

After applying speculative decoding to the target model with the help of a draft model, we were able to observe speedup compared to the baseline as expected. However, there were a few more notable observations. The speedup rate drops below the base line for certain cases, specifically ones with high gamma factors. Recalling that the idea of speculative decoding originated from hardware branch predictors, then a high gamma factor is analogous to a high latency until the branch is resolved. This leads to an increased penalty if the speculated tokens are incorrect. Intuitively, speculations based on mis-speculations are more likely to be incorrect. Furthermore, the optimal gamma between the two model pairs is different as the optimal gamma can only be determined at run-time. Wang et al. [17] have also observed the issue and developed a framework to determine optimal gamma factor at run time. For us, this means that our hardware accelerator needs to be flexible and can adapt to different gamma factors.

### A. *Observation*

Marchisio et al. [9] observed that transformers are often memory-bound due to the amount of parameters that has to be passed through for each inference step. That is to say, relatively little memory communication occurs outside of inference steps.

We observe that only the tokens and logits are accessed during the verification and rollback phase of speculative decoding. The size of the logits is directly proportional to the model's vocabulary size. Which is approximately 100KB for both GPT-2 and OPT, which is $(\gamma + 1) * 100$KB in total. In hardware, we believe we can buffer the logits in an on-chip SRAM due to its relatively small size. Furthermore, verification stage can be easily pipelined due to its sequential nature.

## V. EVALUATION

We evaluate the verification performance with A100, A6000, Intel i7-12700k, HLS implementation from C, and our optimized HADES implementation. We used GPT-2 (124M parameters) as the draft model, and GPT2-XL (1.5B parameters) as the target model. We measured the time it takes for the verification phase and the number of tokens it verifies to calculate verified tokens per second, which will serve as our quantification metrics. Figure 2 shows the verification performance across different computing devices. HADES achieves 6.99x and 7.74x better tokens/sec than A100 and A6000, respectively. We also observe that i7-12700k has similar performance compared to the GPUs. We speculate that it could be caused by the fact that verification is not computationally intensive but rather memory-bounded, incurring incurred kernel launch overhead.

We also evaluated different devices' energy efficiencies. We used profiling tools to measure the real-time power consumption for A6000, A100, i7-12700k, and AMD Vivado tool to estimate the power consumption of C HLS and HADES. The verification tokens/sec/Watt are plotted in Figure 3. Again, HADES shows massive energy efficiency gain, achieving 159.66x and 117.95x better tokens/sec/Watt than A6000, A100, respectively.

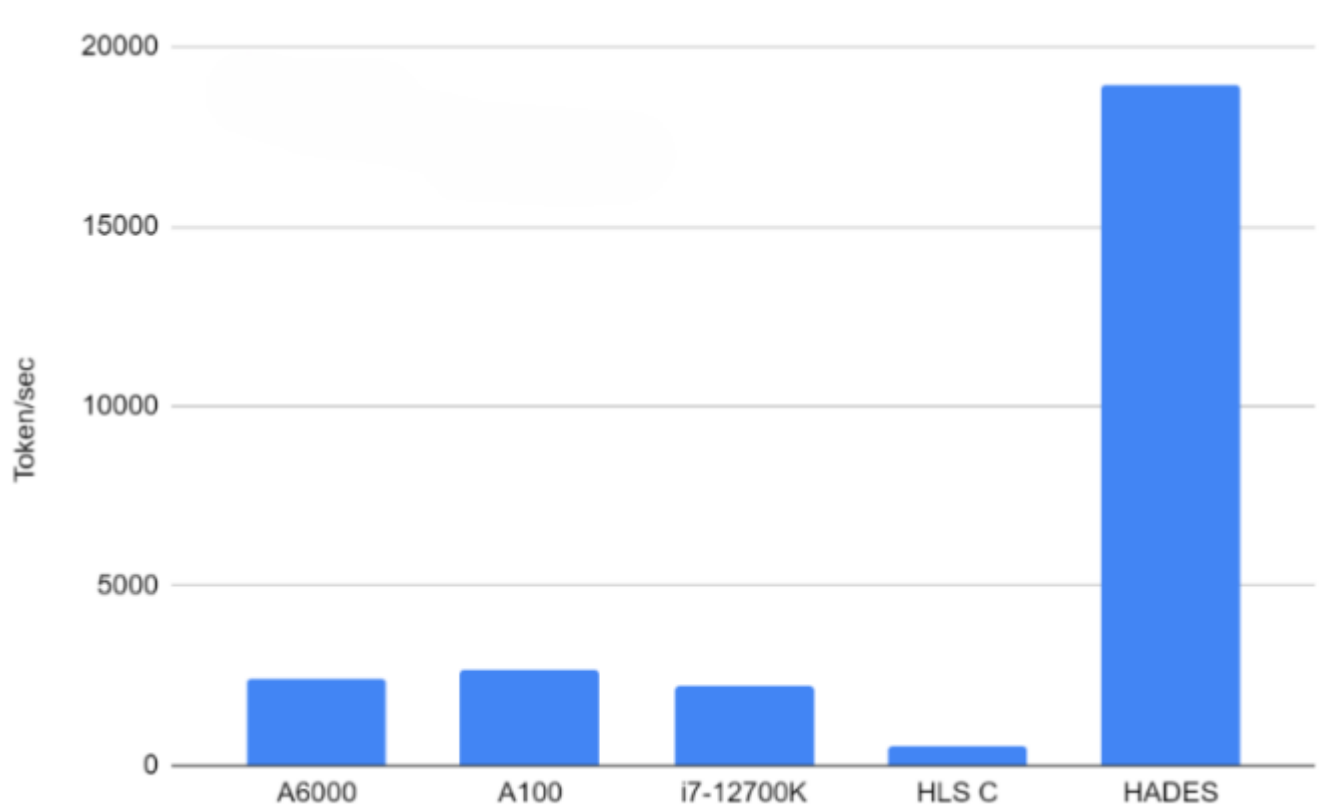

Figure 2. *Verification tokens/sec for RTX-A6000, A100, Intel i7-12700k, HLS implementation from C, and our optimized HADES implementation.*

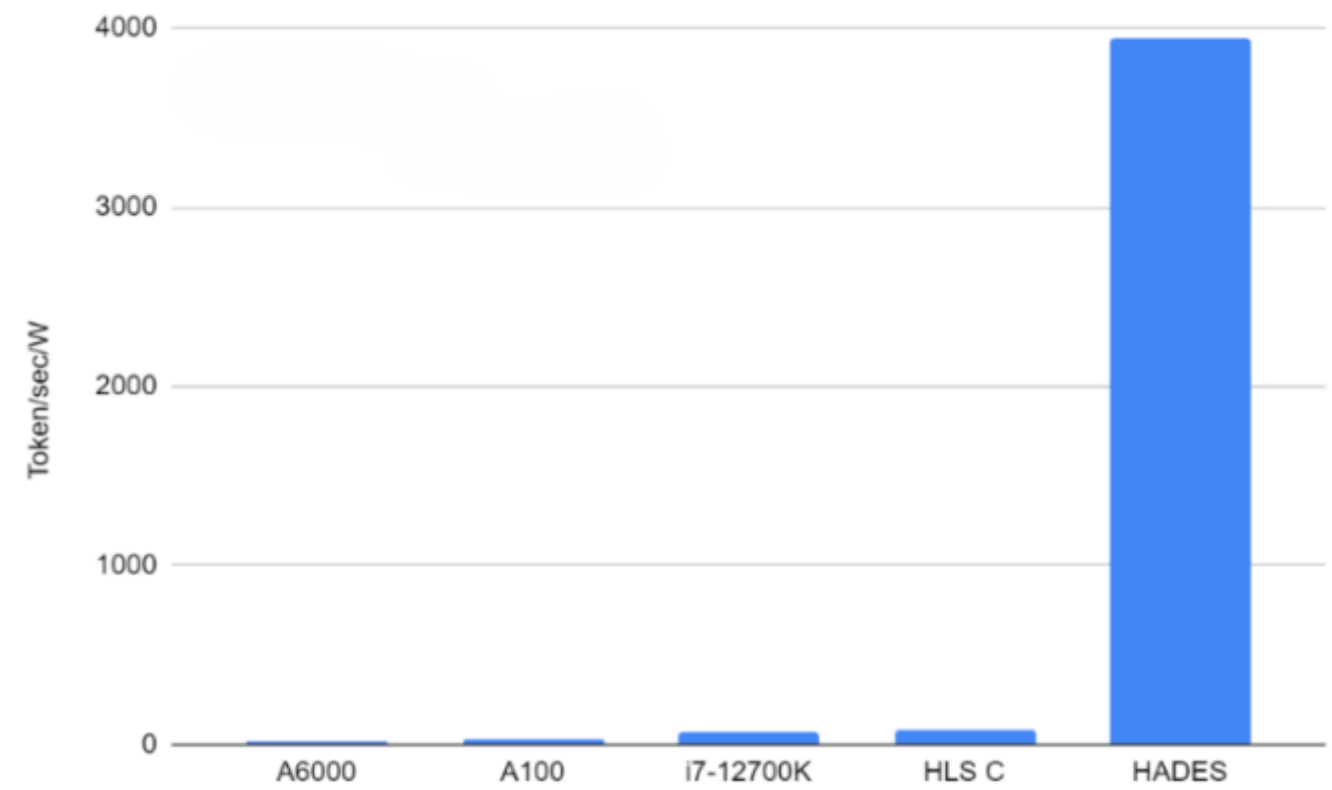

Figure 3. *Verification tokens/sec/Watt for RTX-A6000, A100, Intel i7-12700k, HLS implementation from C, and our optimized HADES implementation.*

## VI. DISCUSSION

**Importance of Specialized Hardware**: One of the most significant takeaways from our project is the tremendous impact that specialized hardware can have on the performance and energy efficiency of LLMs. By designing hardware specifically tailored for speculative decoding, we achieved substantial improvements over general-purpose GPUs. This underscores the potential benefits of continuing to explore and develop specialized hardware solutions for various computational tasks in natural language processing.

**Challenges of End-to-End Implementation**: Implementing end-to-end speculative decoding at the hardware level presented unique challenges. Due to time limitation, we have to only focus on the verification phase first. However, it lowers the impact of our work, since verification phase only occupies about 2% to 10% of end-to-end speculative decoding.

**Energy Efficiency Considerations**: Our work demonstrated significant energy efficiency gains, which are crucial for the sustainability of deploying large-scale models in real-world applications. This aspect of our project emphasized the necessity of considering energy consumption alongside performance metrics, particularly as the scale and complexity of LLMs continue to grow.

**Scalability Issues**: As we benchmarked our accelerator with various models, it became clear that scalability is a critical factor. The ability to maintain performance gains across different model sizes and architectures is essential for the practical adoption of our approach. This experience reinforced the importance of designing scalable solutions that can adapt to evolving model requirements.

**Interdisciplinary Collaboration**: The project benefited greatly from the collaboration between hardware and software experts. This interdisciplinary approach facilitated a more comprehensive understanding of the challenges and opportunities in implementing hardware-accelerated speculative decoding. For instance, software experts identified specific parts of the LLM model, such as the verification stage, that could benefit from hardware parallelization. They worked closely with hardware experts to ensure these parts were clearly explained and aligned with hardware capabilities. Conversely, hardware experts provided insights into the unique features and limitations of the hardware, enabling software experts to adapt their algorithms accordingly. This iterative exchange not only optimized the system's performance but also highlighted the value of cross-domain knowledge and the need for ongoing collaboration between different fields of expertise.

**Future Directions**: Finally, our work has opened several avenues for future research, including benchmarking additional models, and leveraging insights from representative studies on FPGA-based LLM accelerators [22] to develop an end-to-end hardware accelerator tailored for speculative decoding. Additionally, we aim to explore multi-candidate speculative decoding verification.

These directions promise to further enhance the performance and efficiency of LLMs, making them more accessible and practical for a wider range of applications.

In conclusion, the development of our hardware-accelerated speculative decoding system has been a challenging yet rewarding endeavor. The lessons learned from this project will undoubtedly inform and inspire future efforts in the field of LLMs and beyond.

## VII. CONCLUSIONS

In this paper, we introduced the first specialized hardware support for speculative decoding in Large language models (LLMs). Our novel approach significantly enhances the performance and energy efficiency of LLM operations. Specifically, our hardware-accelerated speculative decoding achieves verification speeds that are 6.99 times and 7.74 times faster than those of the A100 and A6000 GPUs, respectively.

Moreover, our approach demonstrates remarkable improvements in energy efficiency, being 117.95 times and 159.66 times more energy-efficient than the A100 and A6000 GPUs during the verification phase. These substantial gains highlight the potential of hardware-accelerated speculative decoding to overcome the computational challenges posed by large-scale LLMs, paving the way for more advanced and efficient applications in natural language processing.